\documentclass[runningheads]{llncs}

\usepackage{eccv}

\usepackage{eccvabbrv}

\usepackage{graphicx}
\usepackage{booktabs}
\usepackage{siunitx}

\usepackage[accsupp]{axessibility}  

\usepackage{hyperref}

\usepackage{orcidlink}
\newcommand{\ocb}{OpenCapBench }

\begin{document}

\title{OpenCapBench: A Benchmark to Bridge Pose Estimation and Biomechanics}


\author{Yoni Gozlan\inst{1} \and
Antoine Falisse \inst{1} \and
Scott Uhlrich \inst{1} \and
Anthony Gatti \inst{1} \and
Michael Black \inst{2} \and
Akshay Chaudhari \inst{1}
}

\authorrunning{Y.~Gozlan \and A.~Falisse \and et al.}

\institute{Stanford University \and
Max Planck Institute for Intelligent Systems}

\maketitle

\begin{abstract}
Pose estimation has promised to impact healthcare by enabling more practical methods to quantify nuances of human movement and biomechanics.
However, despite the inherent connection between pose estimation and biomechanics, these disciplines have largely remained disparate. 
For example, most current pose estimation benchmarks use metrics such as Mean Per Joint Position Error, Percentage of Correct Keypoints, or mean Average Precision to assess performance, without quantifying kinematic and physiological correctness - key aspects for biomechanics. 
To alleviate this challenge, we develop OpenCapBench to offer an easy-to-use unified benchmark to assess common tasks in human pose estimation, evaluated under physiological constraints. OpenCapBench computes consistent kinematic metrics through joints angles provided by an open-source musculoskeletal modeling software (OpenSim). 
Through OpenCapBench, we demonstrate that current pose estimation models use keypoints that are too sparse for accurate biomechanics analysis. To mitigate this challenge, we introduce SynthPose, a new approach that enables finetuning of pre-trained 2D human pose models to predict an arbitrarily denser set of keypoints for accurate kinematic analysis through the use of synthetic data. Incorporating such finetuning on synthetic data of prior models leads to twofold reduced joint angle errors. Moreover, OpenCapBench allows users to benchmark their own developed models on our clinically relevant cohort. 
Overall, OpenCapBench bridges the computer vision and biomechanics communities, aiming to drive simultaneous advances in both areas.


\end{abstract}    
\section{Introduction}
\label{sec:intro}
\renewcommand\labelitemi{--}

\begin{figure}[h]
\centering
\includegraphics[width=\linewidth]{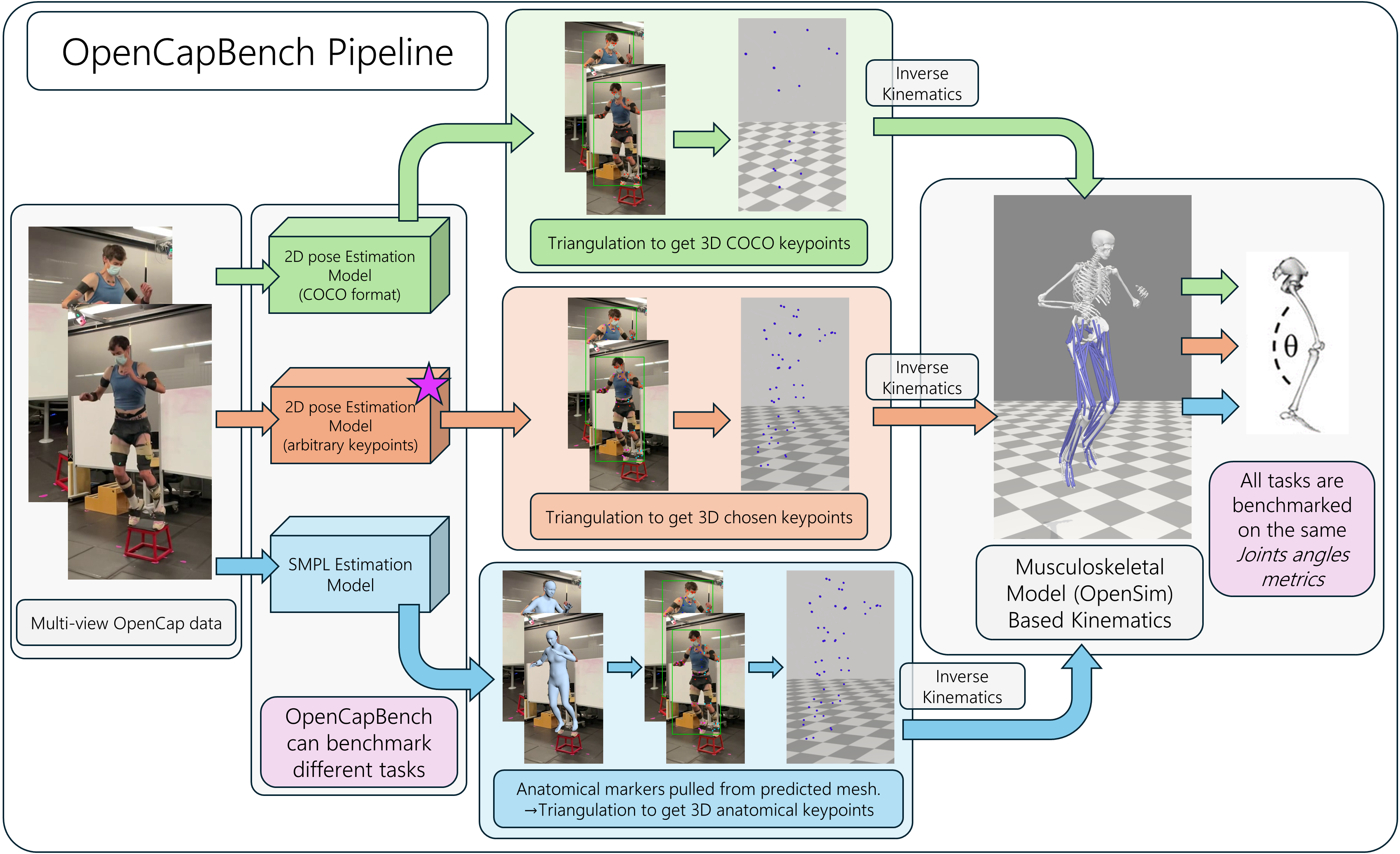}
\caption{\ocb pipeline. SynthPose, our method to finetune 2D pose estimation models to predict any set of body keypoints (designated by a star here) is detailed in \ref{fig:finetuning}.}
\label{fig:main}
\vspace{-5mm}
\end{figure}

A major part of kinematic biomechanical analysis is the study of joint angles that are critical for understanding the interplay between body segments for use in applications ranging from diagnostics \cite{injuries, kine1984, kine2014} and intervention strategies \cite{ResearchMethodsinBiomechanics} to optimizing athletic performance \cite{sportInjuries}. Traditionally, acquiring high-quality kinematic data for research and clinical studies requires a dedicated gait laboratory with synchronized high-speed cameras, application of multiple optical motion markers, and expert personnel trained in biomechanics. These cumbersome requirements make clinical assessment and large-scale clinical trials cost-prohibitive. Meanwhile, in the domain of computer vision, pose estimation models strive to capture and predict human movement from single or multiple videos. Yet, despite the clear intersections in the objectives of the biomechanics and pose estimation domains, there remains a disconnect in their methodologies and evaluations. Importantly, biomechanical models for estimating kinematics constrain joints to move in physiologically realistic ways. For example, while biomechanics researchers constrain movement of the knee to only have one degree-of-freedom joint \cite{knee_dof}, computer vision approaches use physiologically implausible unconstrained three degree-of-freedom motion \cite{SMPL-X:2019}.


Evaluating kinematic metrics is not only important in clinical and sports biomechanics, but can also improve machine learning approaches to pose estimation \cite{kinect}. Using more physiologic joints may serve as a form of regularization, thus improving estimated poses. Furthermore, kinematic metrics provide a richer, temporally consistent, and functionally relevant evaluation compared to traditional pose estimation metrics like Mean Per Joint Position Error (MPJPE) \cite{humans36}, Percentage of Correct Keypoints (PCK) or mean Average Precision (mAP) \cite{poseSurvey}. Kinematic metrics such as Root Mean Squared Error (RMSE) of joint angle better encapsulate the complexities and constraints of human motion, and by emphasizing these metrics, models might generalize more effectively across diverse and out-of-distribution poses, view angles, and occlusions \cite{handKinematics, poseAnatomicalConstraints}. 

Biomechanical studies \cite{OpenCap} show that popular computer vision-based pose estimation models and datasets with sparse keypoint annotations only on joint centers \cite{COCO, OpenPose} result in large errors in joint angles. These errors are likely owing to the fact that only estimating joint centers leaves identifying specific joint angle contributions from the three anatomical axes unconstrained. Therefore, while typical computer vision metrics focused on keypoints accuracy might be satisfactory, specific joint kinematics can still have large errors, illustrating the need for improved benchmarks and metrics of pose estimation. 

Connecting the realms of biomechanical kinematic analysis and pose estimation with computer vision can benefit both fields. Tighter integration can provide real-world benchmarks for computer vision researchers, while translating promising pose estimation models into clinical practice can benefit biomechanics researchers. Against this backdrop, our work aims to bridge the current separation between these disciplines. Our contributions are as follows:

\begin{itemize}
\item We introduce \textbf{OpenCapBench}, a benchmark to align the fields of biomechanics and pose estimation. \ocb includes a \textbf{fully automated pipeline} to streamline the transfer from pose estimation results to the widely-used musculoskeletal modeling and simulation software OpenSim \cite{OpenSim}. This integration allows computer vision experts to seamlessly generate kinematic analyses, without requiring expertise in musculoskeletal biomechanical modeling. 

\item We introduce \textbf{SynthPose, a novel method that uses synthetic data to allow efficient finetuning of pretrained pose estimation models to predict a denser set of keypoints} and improve biomechanical analysis. 
\item Using our new kinematic benchmark, we show that compared to sparse keypoints, our Synthpose method \textbf{twofold reduces average joint angle RMSE} and up to fourfold for certain biomechanically-relevant body joints.
\end{itemize}
The benchmarking pipeline and the different components of SynthPose will be available \hyperlink{https://github.com/yonigozlan/OpenCapBench}{here}.

In uniting the strengths of biomechanics and computer vision, we envision a future where pose estimation models are not just technically proficient, but can also help improve human movement analysis and human health outcomes.

\section{Related Work}
\label{sec:relatedwork}


\textbf{2D Pose Estimation:} 
Datasets like COCO \cite{COCO} and MPII \cite{MPII} include extensive annotated human keypoints for solving monocular pose estimation tasks across a wide array of static images, with annotations for 17 and 16 2D body joint centers, respectively. However, 2D pose estimation datasets do not offer depth information which is crucial for understanding realistic 3D movement patterns.
\\
\\
\textbf{Video-Based Pose Estimation:} Temporal datasets such as JHMDB \cite{JHMDB} and PoseTrack \cite{posetrack} have introduced the challenge of maintaining consistency across frames, a step towards measuring continuous motion pertinent to biomechanical research. However, current computer vision benchmarks primarily focus on visual consistency rather than biomechanical accuracy, indicating a gap for new benchmarks, metrics, and datasets that also evaluate temporal kinematic consistency. 
\\
\\
\textbf{3D Pose Estimation:}
Datasets such as Human3.6M \cite{humans36} and CMU Panoptic \cite{cmupanoptic} have enabled 3D monocular and multi-view pose estimation methods. However, even translating monocular 3D poses into biomechanically valid models remains a complex task \cite{moncularKinematics}. Current pose estimation metrics primarily focus on joint position accuracy and fail to assess biomechanical factors such as joint angle limits. This leads to the generation of physiologically implausible movement solutions that are inadequate for comprehensive biomechanical analysis. 
Multi-view 3D pose estimation allows for more precise 3D reconstructions, especially in datasets with occlusions, which often hinder monocular estimations. Yet, there exists no rigorous benchmark specifically designed to evaluate the biomechanical plausibility of these estimations.
\\
\\
\textbf{Motion Capture Datasets:} Conventional motion capture (MoCap) technology is a specialized and costly resource that requires multiple hours for data collection per subject. Yet, it is indispensable for producing high-fidelity human movement data. MoCap has given rise to datasets such as OpenCap \cite{OpenCap}, MoVi \cite{movi} or PSU-TMM100 \cite{psutmm100}, which offer detailed continuous kinematic human motion data. While MoCap provides unparalleled accuracy, its reliance on elaborate equipment and controlled environments limits its accessibility and scalability. Despite these limitations, MoCap datasets enable validating and testing pose estimation models, to ensure that the models trained on more generalized data can be benchmarked against the "gold standard" of human movement data, as well as providing the necessary ground truth for benchmarking biomechanically grounded kinematic metrics such as joint angles \cite{joints_angles}.
\\
\\
 \textbf{Biomechanical Validity and Parametric Models:} The development of the SMPL model \cite{SMPL:2015} and its successor SMPL-X \cite{SMPL-X:2019} has been instrumental in introducing the notion of parametric body shape. These models have facilitated the creation of datasets used for 3D body shape and pose estimation \cite{rich, 3DPW} with annotations based on the SMPL framework. Although useful, even such data often lack precision necessary for biomechanically accurate pose estimation. The pursuit of biomechanical validity has thus seen the adoption of models from biomechanical research like the Rajagopal et al. model \cite{Rajagopal}, which is implemented in the OpenSim musculoskeletal modeling platform \cite{OpenSim}. The recent introduction of the SKEL model \cite{SKEL}, a SMPL-like model with more biomechanically-realistic degrees of freedom represents another step towards biomechanically accurate modeling of human movement. 
\\
\\
\textbf{Synthetic Data for Pose Estimation:} Synthetic data has recently found applications in a wide range of fields, including image classification\cite{ICsynth}, natural language processing \cite{NLPsynth}, healthcare \cite{healthsynth} and more. The field of pose estimation is no exception: synthetic datasets using SMPL to model and animate their subjects \cite{BEDLAM, AGORA, SURREAL, InfiniteForm} seek to overcome the lack of labeled multi-view 3D data. Synthetic data enables sampling a wide array of human poses and shapes in diverse environments, and provides rich annotations of different types such as 2D/3D keypoints, SMPL meshes, segmentation masks etc. However, synthetic data still faces the same challenge as real data; they require better evaluation metrics to assess biomechanical outcomes.
\\
\\
\textbf{Computer Vision for Biomechanical Metrics Estimation:} Recent efforts have been channeled into leveraging computer vision techniques to estimate biomechanical metrics directly \cite{single_cam_mov_analysis}. The OpenCap \cite{OpenCap} platform uses pose estimation outputs for accurate biomechanical analysis in two stages - first by predicting sparse keypoints using pose estimation models followed by lifting from a sparser to denser set of keypoints using MoCap data and recurrent networks. Bittner et al explore the challenge of reconstructing 3D kinematics from monocular video data \cite{moncularKinematics}. BioPose-3D \cite{biopose} aims to predict 3D biomechanical joint corrections for video-based joint detection methods. These methodologies underscore a growing trend in leveraging computer vision for biomechanical assessment. Our \ocb framework seeks to extend these efforts by providing a holistic, easily accessible evaluation that benchmarks pose estimation models against biomechanically relevant metrics. 

\section{\ocb Pipeline}


\ocb introduces a comprehensive benchmarking pipeline, designed to evaluate the efficacy of pose estimation models in the context of biomechanics. 

\subsection{Dataset Integration}
The foundation of our benchmark is the OpenCap \cite{OpenCap} dataset, a biomechanically-focused MoCap and multi-view data collection. This dataset includes exercises commonly used in biomechanical studies, such as squats, sit-to-stand, drop jumps and walking, performed by 10 different subjects. There are 16 movements per person, with durations between 2 to 8 seconds. Each movement is recorded by 5 synchronized and calibrated cameras; we include two cameras in our experiments following results from Uhlrich et al that suggest that two cameras provide comparable accuracy to more cameras \cite{OpenCap}. 
OpenCap includes 3D marker data obtained with an eight-camera MoCap system (Motion Analysis Corp., Santa Rosa, CA, USA) that tracked the positions (100 Hz) of 31 retroreflective markers placed on established anatomical landmarks and 20 tracking markers \cite{OpenCap}. The joint angles we use as ground-truth were obtained from the MoCap markers using OpenSim’s Inverse Kinematics tool \cite{OpenSim} and the Rajagopal model \cite{Rajagopal}. 
We also added bounding boxes obtained with high-performing human detection model \cite{convnext} to the OpenCap videos for subject cropping, in order to provide a fair basis to benchmark all pose estimation models on.

\subsection{Benchmarking Pipeline}
Our benchmark is characterized by its versatility and modular design, capable of evaluating a wide range of pose estimation tasks. For the purpose of this paper, we focus on evaluation of 2D single-frame pose estimation models and single-frame SMPL pose and shape estimation models.
We apply these models on the video sequences from OpenCap Cam1 and Cam3, adopting a common pipeline to go from 2D single frame pose estimation to 3D joints kinematics detailed below. The goal of this common pipeline is to provide an automated and fair baseline to evaluate the most basic tasks in pose estimation while losing as little information as possible in the process. 
However, we note that the way \ocb is designed allows adapting to additional tasks, such as 3D multi-view , 3D monocular, 2D/3D temporal pose estimation or even direct kinematics estimation, by selectively bypassing or modifying components in our modular pipeline.

\subsubsection{2D keypoints extraction}
\begin{figure}
    \centering
    \begin{minipage}{0.43\textwidth}
          \centering
          \begin{minipage}[b]{0.45\linewidth}
            \includegraphics[width=\linewidth]{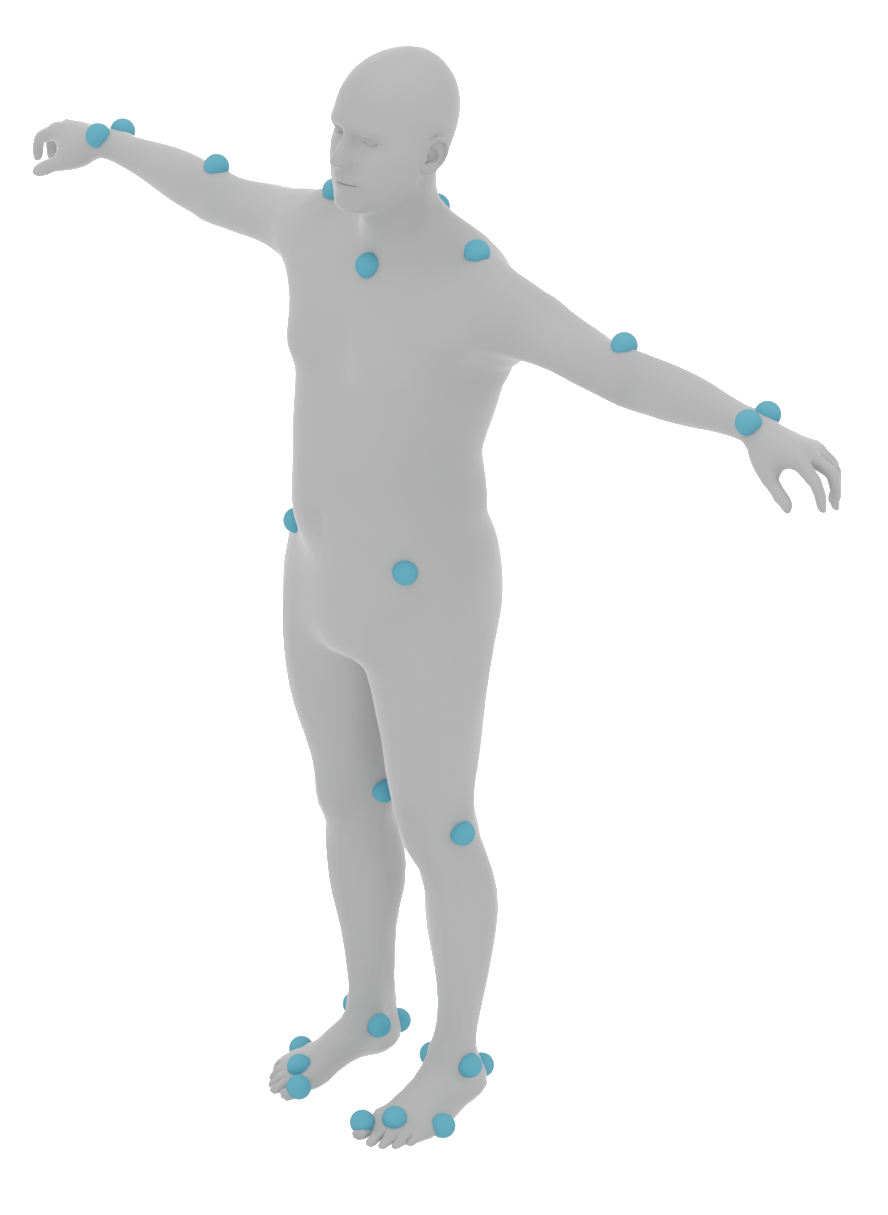}
          \end{minipage}
          \begin{minipage}[b]{0.45\linewidth}
            \includegraphics[width=\linewidth]{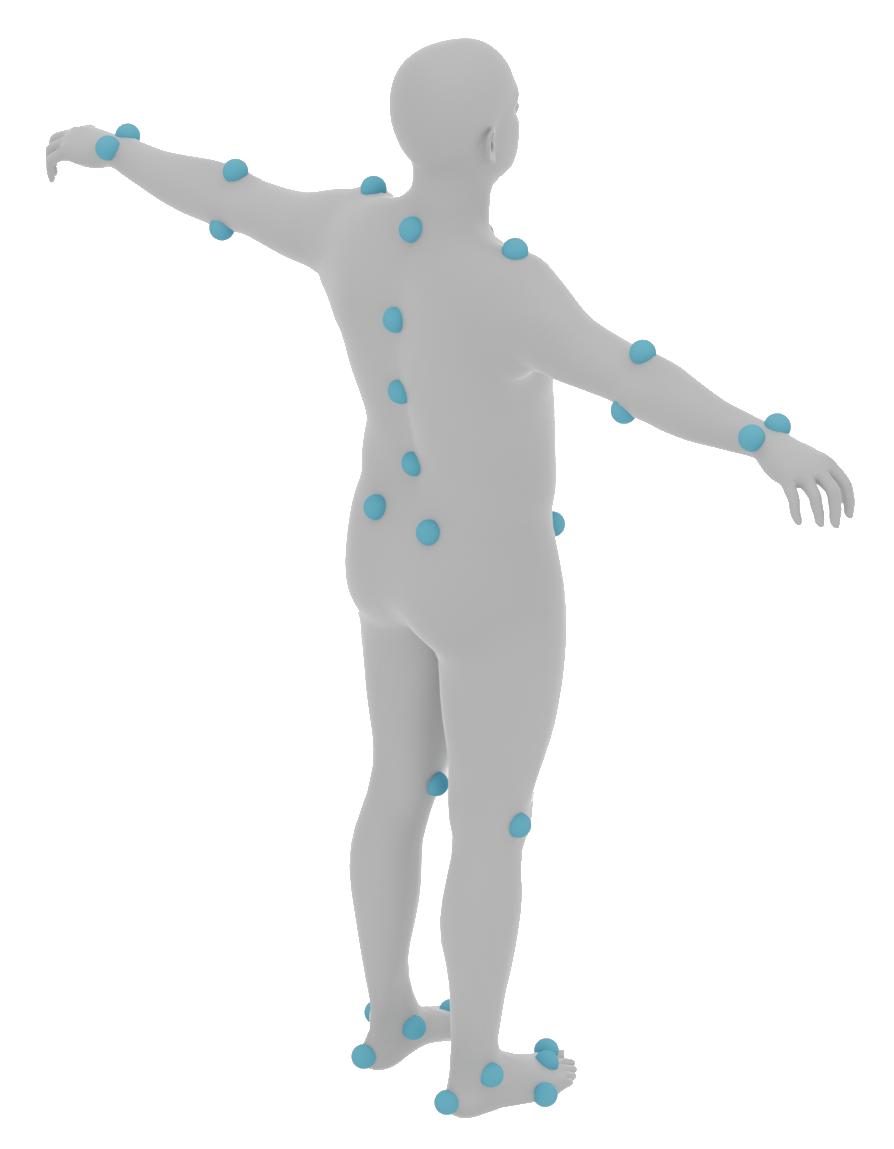}
          \end{minipage}
        \caption{Chosen subset of 35 vertices \\ from SMPL mesh.}
        \label{figure:sub_vertices}
    \end{minipage}
    \begin{minipage}{0.56\textwidth}
        \centering
        \includegraphics[width=\textwidth]{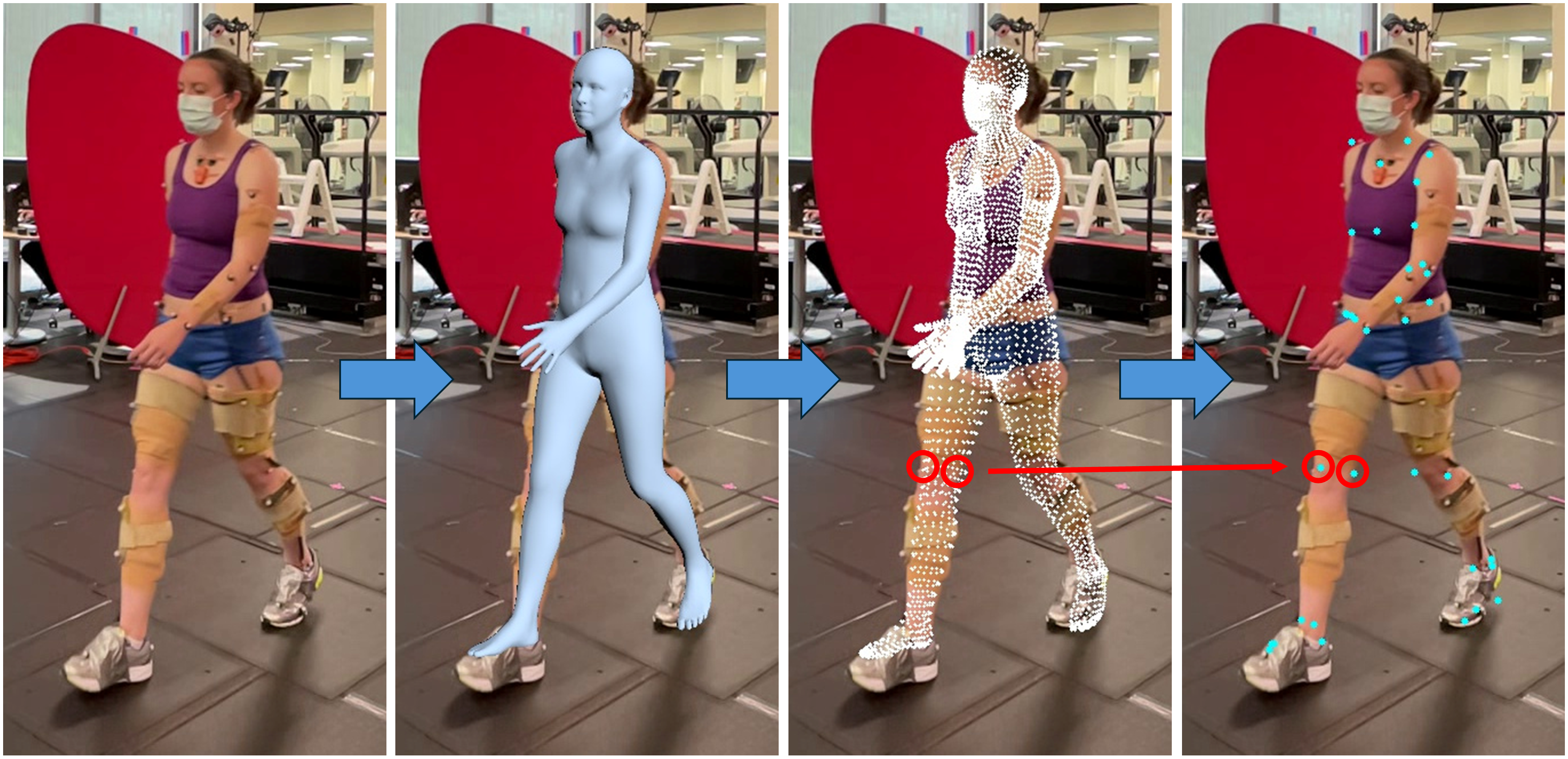} 
        \caption{Extracting 2D keypoints from SMPL mesh.}
        \label{fig:projection_pred}
    \end{minipage}
\vspace{-3mm}
\end{figure}

The first stage of the \ocb pipeline \ref{fig:main} involves extracting a set of 2D body keypoints for each frame of the two different camera feeds.
\textbf{This is where users wanting to test their single frame 2D pose or SMPL shape estimation models can easily integrate their models.}
2D keypoint extraction can be performed in one of two ways:
\begin{itemize}
    \item \textbf{Using a 2D pose estimation model predicting any set of 2D body keypoints} (as long as the Inverse Kinematics setup has been defined for this set). The only post-processing done to the outputted predictions is temporal denoising prior to multi-view triangulation using a dual pass low-pass smoothing with a Butterworth filter with a cut-off frequency of 30Hz. We have integrated the MMPose framework \cite{mmpose2020} 2D pose inference pipeline to \ocb, thus if a model is available on MMPose, it can directly be benchmarked on \ocb without further modification needed. Otherwise, the user can plug-in their model inference function to the pipeline as described in our Github repository.
    \item \textbf{Using a SMPL shape estimation model.} Users can plug-in their SMPL shape estimation model inference function to the pipeline as described in our Github repository. This time, the post-processing consists in projecting a subset of vertices from the predicted SMPL shape onto the image, as illustrated in \ref{fig:projection_pred}. By default, the projected set of vertices corresponds to a subset of anatomical markers used in MoCap setups \cite{OpenCap} and illustrated in \ref{figure:sub_vertices}. This set of marker was manually chosen by experienced biomechanics researchers using SMPL Blender add-on, by selecting the anatomically closest SMPL vertex for each anatomical marker. The anatomical markers are derived specifically to create biomechanically relevant 3D joint coordinate
systems per segment, based on the recommendations of the International Society for
Biomechanics \cite{ISB} .The same post-processing step as for 2D pose estimation is applied to the extracted keypoints. Of course, due to the modular nature of \ocb, other marker sets can be used. We include a SKEL tool \cite{SKEL} to either use the default set of markers suggested by SKEL or alternatively, users can define a personalized set and regress the corresponding OpenSim musculoskeletal model markers using the visualization tool.
\end{itemize}

\subsubsection{3D Triangulation}
Following 2D estimation, the framework employs a deterministic triangulation algorithm to combine multi-view 2D keypoints into 3D keypoints with real-world absolute distances, using viewpoints from two calibrated cameras (Cam1 and Cam3 as defined in OpenCap \cite{OpenCap}).

\subsubsection{Inverse Kinematics (IK)}
 The IK step uses the Rajagopal \cite{Rajagopal} musculoskeletal model to estimate joint kinematics from a sequence of 3D keypoints. This step is performed through OpenSim's python API. Our framework is versatile, providing IK configuration files for common landmark sets (such as COCO, OpenPose, COCO whole body) as well as for the subset of MoCap markers described in \ref{figure:sub_vertices}. Again, we include a SKEL tool \cite{SKEL} which allows the use of a personalized set of SMPL markers and regress the corresponding markers on the OpenSim model.
 


\subsection{Evaluation Metrics}

From the outputted joint kinematics, we use a subset of joint angles to be compared with the ground truth joint angles obtained with the MoCap setup of OpenCap \cite{OpenCap}. We use joint angles for \textit{Pelvic Tilt, Pelvic List, Pelvic Rotation, Hip Flexion, Hip Adduction, Hip Rotation, Knee Flexion, Ankle Flexion, Subtalar Inversion/Eversion, Lumbar Extension, Lumbar Bending, and Lumbar Rotation}, which correspond to lower-body kinematics as they are the focus of OpenCap and guaranteed to have accurate ground-truths.
Following the literature \cite{kineAcc, kineIMU, OpenCap}, we use RMSE for the entire waveform for each joint and each trial as metrics.

\subsection{Leaderboard}
\ocb will feature different leaderboards for individual tasks alongside a global leaderboard encompassing all tasks. This initiative aims to cultivate competition, motivating the computer vision community to develop kinematically accurate pose estimation models, and serve as a resource for biomechanists seeking optimal camera-based kinematics prediction methods.
The leaderboard will include all joint angles metrics specified above separately, as well as an average of those to establish a ranking.

\section{Arbitrary 2D keypoints estimation using synthetic data}
\label{sec:methods}
\subsection{Motivation}

Previous biomechanics studies \cite{OpenCap, 3DjointsAcc} show the shortcomings of typical pose estimation models which predict sparse sets of keypoints in format such as COCO, OpenPose or MPII. Not only are these models trained and evaluated on manually annotated data, which do not guarantee precise annotations., but the sparse number of keypoints they predict (17-22 keypoints) do not fully characterize the translations and rotations of all body segments. This inadequacy is accentuated between the hips and the shoulders due to the lack of keypoints in this area \cite{OpenCap}. Using this limited marker set for inverse kinematics is thus susceptible to result in large angular joint errors.

OpenCap findings \cite{OpenCap} show that kinematic metrics can be drastically improved using an LSTM augmenter trained to predict a time series of anatomical markers (corresponding to a subset of MoCap markers) from a time series of sparse keypoints (in COCO-like format). However, this approach is based on MoCap data limited in diversity and captured in controlled environments, and may be prone to overfitting and imprecision since the original image priors are lost in this process. Thus we hypothesize that predicting a denser set of anatomically meaningful keypoints directly from images, akin to MoCap setups, will improve joint angle metrics after inverse kinematics.

To obtain such a model, we can leverage the characteristics of the SMPL model that maintain a fixed mesh topology, meaning that the number of vertices and their connectivity (i.e., the mesh structure) remains unchanged independent of the body shape and pose parameters. This allows for the identification of vertices corresponding to specific anatomical features on the SMPL mesh, such as subsets of MoCap landmarks.

Thus, one potential solution to automatically predict new keypoints involves utilizing an existing model capable of predicting SMPL parameters (or mesh directly \cite{virtualMarkers}) from an image, generating a SMPL mesh from these parameters, extracting a pre-defined subset of SMPL vertices, and projecting them onto the original images, as described in the previous section and in figure \ref{fig:projection_pred}. However, this approach presents computational challenges as predicting SMPL parameters and generating meshes are resource-intensive tasks. Moreover, the current state-of-the-art SMPL prediction models do not offer the desired level of accuracy compared to landmark estimation models.

\subsection{Leveraging synthetic data to finetune pose estimation models}

To address these challenges, we introduce SynthPose, a novel approach for training pose estimation models to predict an arbitrary subset of body keypoints derived from SMPL mesh vertices.

In this work, we focus on predicting bony anatomical keypoints to maximize kinematic accuracy. However, this method can be adapted to specialize models for predicting any sets of body keypoints, such as hands, feet, or head keypoints.
\begin{figure}[h]
\centering
\includegraphics[width=\linewidth]{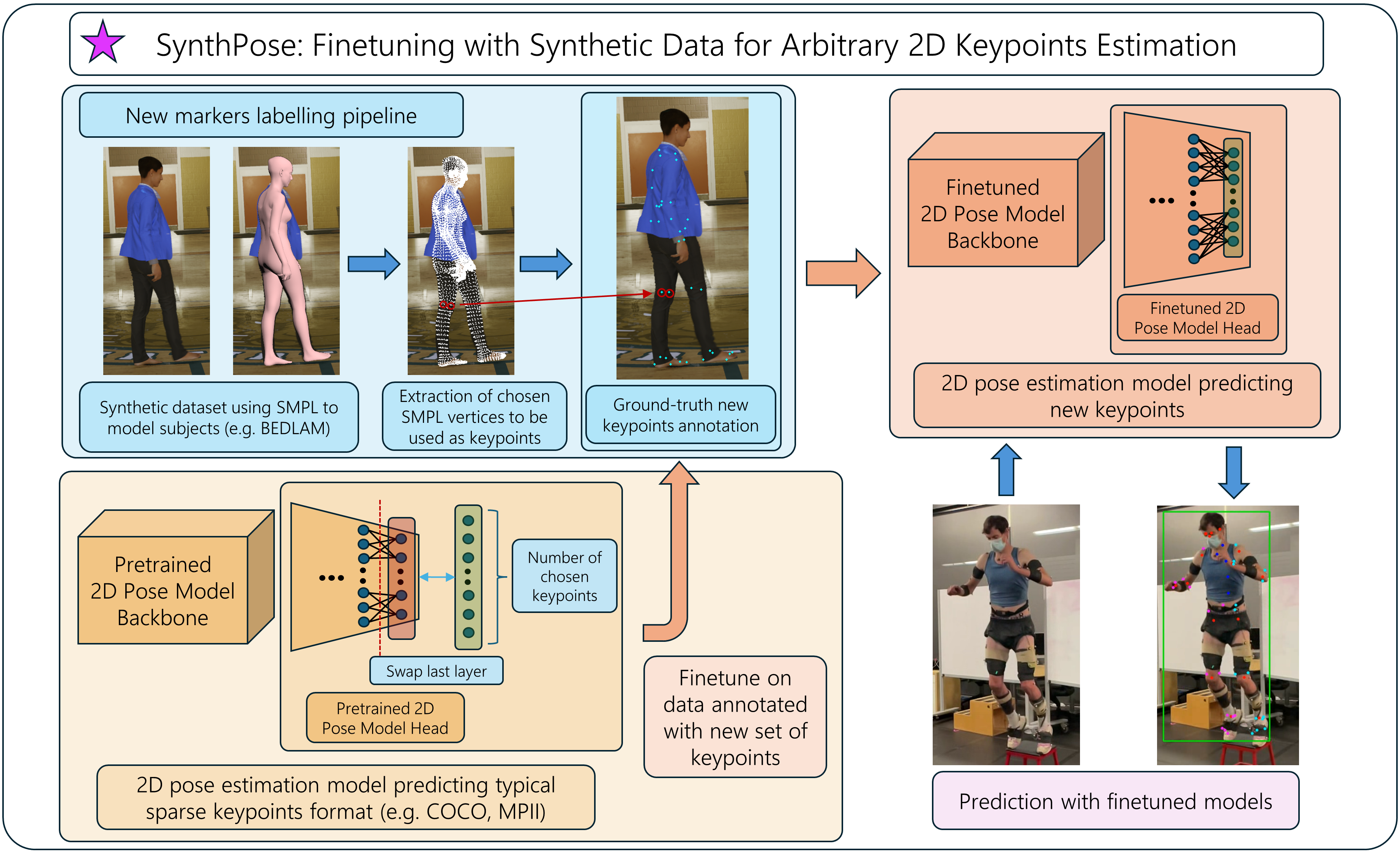}
\caption{SynthPose: a new method leveraging finetuning of pose estimation models on synthetic data to predict an arbitrary set of 2D keypoints.}
\label{fig:finetuning}
\vspace{-3mm}
\end{figure}
\\
\\
\textbf{Labelling pipeline}. As illustrated in \ref{fig:finetuning}, a major component of our new method is an automated labelling pipeline, which can create 2D keypoints annotations of a chosen subset of SMPL vertices on any synthetic dataset that uses SMPL to model its subjects. To do so, we generate the SMPL meshes corresponding to the subjects in the synthetic data using the SMPL pose and shape parameters given in the dataset annotations, project them onto the images and extract the specified subset of vertices to be used as 2D keypoints annotations.
The projected keypoints $P_{2D}$ depend on the given camera's intrinsic and extrinsic matrices $K$ and $V$ respectively, and are computed as:
\[ P_{2D} = K \times E \times V_{smpl} \]
Where \( V_{spml} \) denotes the subset of vertices of the SMPL model.
\\
\\
\textbf{Finetuning.} The second major component of SynthPose is a finetuning process, where we take a pretrained 2D pose estimation models predicting typical set of keypoints (COCO, MPII etc.), swap the last layer for a layer adapted to the new set of keypoints we want to predict, and finetune this model on our synthetic labelled dataset.
\\
\\
Leveraging synthetic datasets offers several advantages.
Besides the benefits of the SMPL model detailed above, these synthetic datasets offer exact annotations, as the mesh of the 3D models depicted in the different images of the datasets are precisely the SMPL meshes. This contrasts with datasets derived from real-world data subsequently annotated with SMPL models, which often exhibit shortcomings in annotation precision, because of the difficulty in precisely manually aligning a SMPL mesh on a 2D image.

By design, SynthPose can leverage learned features from traditional pose estimation training, by using weigths from models trained on datasets like COCO or MPII as initial weights for the new models backbone and prediction head. This transfer learning approach not only significantly reduces training time but also enhances model performance. 

In our study, we carefully selected datasets for fine-tuning our models designed to predict arbitrary keypoints. Each dataset was chosen with specific considerations in mind:
\\
\\
\textbf{BEDLAM Dataset \cite{BEDLAM}:} This extensive synthetic dataset provides a large-scale foundation for training our models.
\\
\\
\textbf{Infinity VisionFit \cite{Infinity, InfiniteForm}:} This synthetic dataset, generated using the Infinity VisionFit API (Infinity AI), includes out-of-distribution samples featuring individuals engaged in various exercise routines. This movement diversity is beneficial in enabling our models to generalize effectively across different scenarios.
\\
\\
\textbf{3DPW Dataset \cite{3DPW}:} To address the limitations of our training set which lacks real data and subjects wearing shoes, we incorporate the 3DPW dataset. This dataset contains in-the-wild data of 18 subjects in 60 different scenes. Despite imperfections in annotations, this dataset supplements our training data with valuable real-world examples to help close the sim-to-real gap.
\\
\\
\textbf{COCO Dataset \cite{COCO}:} In addition to the selected SMPL vertices, our models are designed to also output keypoints in the COCO format. Therefore, we integrate the COCO dataset into our aggregated training dataset. This inclusion also helps mitigate the sim-to-real distribution shift and thus, prevent catastrophic forgetting.
\\
\\
In the next section, we demonstrate that our proposed method significantly outperforms the approach of using state-of-the-art SMPL shape and pose estimation models, followed by projecting a subset of the predicted mesh vertices to predict arbitrary keypoints.

\section{Experiments}
\label{sec:experiments}
\newcommand{\ra}[1]{\renewcommand{\arraystretch}{#1}}
\subsection{New arbitrary 2D keypoints prediction method}

In this section, we present experiments which showcase the benefit of using our new method to predict arbitrary 2D keypoints compared to state-of-the-art SMPL estimation models.
We will first compare our results on a conventional 2D keypoints estimation benchmark using Percentage of Correct Keypoints (PCK) normalized by bounding boxes at different levels of precision, and then use \ocb for comparison.
For this experiment, we choose a subset of 35 SMPL vertices corresponding to a subset of anatomical markers typically used in MoCap setups \cite{Rajagopal, OpenCap}, detailed in Figure \ref{figure:sub_vertices}. 
We study several state-of-the-art models for each task, namely two SMPL estimation models, CLIFF \cite{cliff} and VirtualMarkers \cite{virtualMarkers}, and two 2D pose estimation models, HRNet-W48+DARK \cite{hrnetDark} (referred as HRNet-W48 in the tables) and VitPose (Base and Huge) \cite{vitpose}. We choose these 2D pose estimation models to
represent diverse SOTA model types (CNN and ViT) and determine potential architecture-based biases. We evaluated
two ViT sizes to show that results are coherent with expectations (i.e., Huge outperforms Base).\

\subsubsection{Results on RICH}


We first benchmark different models on the RICH dataset \cite{rich}. The RICH dataset captures multi-view outdoor and indoor video sequences of diverse subjects performing different physical activities. While not attaining the same level of precision as synthetic data annotations, the provided SMPL annotations that we use to extract ground-truth 2D keypoints (as in \ref{projection_pred}) represent the current best achievable quality in datasets containing real-world data. This high-quality data is a result of SMPL meshes being fitted to 3D human bodies captured through markerless motion capture and 3D body scans. RICH also includes high resolution 3D scene scans which allows for accurate vertex-level contact labels on the body. Therefore, we have deemed this dataset the ideal choice for evaluating our novel arbitrary keypoint prediction technique. 



\begin{table}[ht]
\centering
\scalebox{1}{
\ra{0.8}
\setlength\tabcolsep{15pt}
\begin{tabular}{@{}lccc@{}}\toprule
& \multicolumn{3}{c}{PCK Precision(\textbf{$\uparrow$})}\\ \cmidrule{2-4}\\
& @0.05 & @0.10 & @0.20\\ \midrule
\multicolumn{4}{l}{\textit{SMPL (MoCap markers extracted)}} \vspace{2pt}\\
\phantom{ab} CLIFF \cite{cliff} & 0.640 & 0.802 & 0.905 \\
\phantom{ab} Virtual Markers \cite{virtualMarkers} &  0.707 &  0.844 & 0.926 \vspace{4pt}\\
\multicolumn{4}{l}{\textit{SynthPose (predicting MoCap markers)}}\vspace{2pt}\\
\phantom{ab} HRNet-W48 \cite{hrnetDark} & 0.892 & 0.958 & 0.982\\
\phantom{ab} ViTPose-B \cite{vitpose} & 0.859 & 0.941 & 0.971\\
\phantom{ab} ViTPose-H \cite{vitpose} & \textbf{0.903} & \textbf{0.966} & \textbf{0.985}\\
\bottomrule
\end{tabular}
}
\vspace{2mm}
\caption{Comparison with SOTA SMPL Models on the RICH \cite{rich} test set. SynthPose significantly outperforms SMPL estimation based methods. Note that CLIFF, VirtualMarkers and HRNet-W48+DARK share the same HRNet backbone architecture, showing the advantage of Synthpose with similarly-sized models.}
\label{table:ablation_results}
\vspace{-5mm}
\end{table}

\begin{table*}[t]
\centering
    \scalebox{0.8}{
\ra{0.8}
\setlength\tabcolsep{3pt}
\begin{tabular}{@{}lcccccccccccc@{}}\toprule
\multicolumn{0}{l}{Joint angle RMSE (\textbf{$\downarrow$})} & \multicolumn{3}{c}{Pelvis} & \multicolumn{3}{c}{Hip} & Knee & Ankle & Subtalar & \multicolumn{3}{c}{Lumbar} \\ \cmidrule{2-4} \cmidrule{5-7} \cmidrule{11-13}
\multicolumn{0}{l}{} & Tilt & Rotation & List & Flex & Add & Rot & & & & Ext & Bend & Rot \\ \midrule
\multicolumn{12}{l}{\textit{2D pose methods (COCO-wholebody keypoints)}}\vspace{1pt}\\
\phantom{ab} HRNet-W48 & 24.8 & 3.7 & 4 & 23.4 & 5.1 & 9.9 & 9.4 & 8.2 & 14.2 & 39.1 & 5.6 & 6\\
\phantom{ab} RTMW-X & 14.4 & 3.5 & 4.5 & 15 & 5.2 & 7.9 & 9.3 & 7.8 & 11.4 & 20.3 & 5.6 & 5.9
\vspace{3pt}
\\
\multicolumn{12}{l}{\textit{SMPL-based models (MoCap landmarks)}}\vspace{1pt}\\
\phantom{ab} CLIFF & \textcolor{Green}{6.1} & 4.7 & 4.7 & 10.5 & 6.7 & 7.4 & 11.9 & 11.8 & 13.2 & \textcolor{Green}{6.5} & 6.3 & 13.3\\
\phantom{ab} Virtual Markers & \textcolor{Green}{6.0} & 4.2 & 4.9 & 8.7 & 5.5 & 8.1 & 9.5 & 11.9 & 10.9 & \textcolor{Green}{8.5} & 6.3 & 9.7
\vspace{3pt}
\\
\multicolumn{12}{l}{\textit{\textbf{Our work:} SynthPose (MoCap landmarks)}}\vspace{1pt}\\
\phantom{ab} HRNet-W48 & \textcolor{Green}{5.9} & 2.9 & \textcolor{Green}{3.5} & 8.9 & \textbf{4.9} & 7.3 & 8.9 & 8.7 & 9.5 & \textcolor{Green}{7.5} & 4.7 & 8.3\\
\phantom{ab} ViTPose-B  & \textcolor{Green}{5.2} & 3.3 & 3.6 & \textbf{8.3} & 5.3 & 8 & 8.6 & 8.7 & 9.3 & \textcolor{Green}{6.6} & 5.1 & 10\\
\phantom{ab} ViTPose-H & \textcolor{Green}{5.1}& \textbf{2.8} & \textcolor{Green}{3.4} & \textbf{8.3} & \textbf{4.9} & 7.3 & 8.3 & 7.6 & \textbf{9.1} & \textcolor{Green}{6.2} & \textbf{4.6} & 8.7\\
\midrule

\multicolumn{12}{l}{\textit{\textbf{Our work adjusted:} SynthPose (MoCap landmarks)}}\vspace{1pt}\\
\phantom{ab} HRNet-W48 & \textcolor{Green}{\textbf{4.3}} & 3.4 & \textcolor{Green}{2.8} & \textbf{8.3} & 6.1 & 7 & \textbf{7.7} & 8.9 & 9.8 & \textcolor{Green}{\textbf{4.5}} & 4.8 & 5.8\\
\phantom{ab} ViTPose-H & \textcolor{Green}{\textbf{4.3}}& 3.5 & \textcolor{Green}{\textbf{2.7}} & 8.6 & 5 & 6.1 & 8 & \textbf{7.5} & 9.3 & \textcolor{Green}{4.8} & \textbf{4.6} & \textbf{5.5}\\
\midrule
\multicolumn{13}{l}{\textit{OpenCap method (using LSTM augmenter trained on 108 hours of MoCap data) \cite{OpenCap}}}
\vspace{2pt}\\
\phantom{ab} HRNet-W48 & 7.4& 2.4 & 3.5 & 6.8 & 3.2 & 4.7 & 4.3 & 6.2 & 6.7 & 9.2 & 3.4 & 5\\
\bottomrule
\end{tabular}
}
\vspace{2mm}

\caption{Cross-Comparison of Results. SynthPose significantly improves over computing kinematics from only COCO keypoints, and outperforms SMPL estimation based 2D MoCap landmarks predictions on all joint angles metrics. We emphasize in green results improving over the method used in OpenCap \cite{OpenCap}, which uses an LSTM augmenter trained on 108 hours of MoCap data.}
\label{table:abitraty_kpts_ocb}
\vspace{-5mm}
\end{table*}


We outperform SOTA SMPL mesh prediction based models by 26\% in PCK at 0.05, showing the clear advantage of our method for this particular task.

\subsubsection{Results on OpenCapBench}

Besides traditional computer vision benchmarking, we benchmark our method on \ocb using RMSE of kinematic joint angles obtained from OpenSim for each individual trial. We also add 2D pose estimation models which predict COCO-wholebody subset of keypoints (from which we don't use the hands or the face landmarks) to the comparison.\\
The results are summarized in table \ref{table:abitraty_kpts_ocb}.

The results illustrate the advantage of our method. Indeed, we observe that keypoints based methods obtain better joint angles prediction overall compared to models predicting SMPL meshes. We also show that predicting a MoCap subset of landmarks over COCO keypoints enables clear improvement on all metrics except Lumbar Rotation, with 3-5x reduced RMSEs for Pelvis Tilt, Hip Flexion and Lumbar Extension metrics.

We underline the fact that we are using the same set of keypoints (described in \ref{figure:sub_vertices}) to compute inverse kinematics on OpenSim for both SMPL-based method and our newly introduced method in these specific experiments. However, tools such as the one introduced with SKEL \cite{SKEL}  allow users to specify any set of SMPL vertices they want to use for inverse kinematics, and visually regress the corresponding markers on the OpenSim model. The set of anatomical keypoints we use is much sparser than the one used by default in SKEL to perform inverse kinematics, which may provide an unfair edge to our method. We simply illustrate here that with a same set of predicted SMPL vertices, our method performs better than the SMPL-based method. However, we encourage the community to test the denser keypoints suggested by SKEL to determine their effect on performance in the SMPL leaderboard of \ocb.

Our proposed synthetic data approach challenges the method used in OpenCap on several metrics, namely Pelvis Tilt, Pelvis Rotation and Lumbar Extension.  Importantly, we only low pass filter raw pose prediction results for our inverse kinematics input. In contrast, OpenCap performs post-processing on the predicted keypoints based on the predictions’ confidence and use an LSTM keypoints augmenter model which was trained on 108 hours motion capture data \cite{OpenCap}. This augmenter converts the predicted 3D keypoints from COCO or OpenPose format to MoCap keypoints, leveraging temporal priors in marker prediction.

\subsection{Benefits of OpenCapBench}
Here, we aim to illustrate how OpenCapBench can offer insights on models that current benchmarks and metrics cannot. To show this, we propose an ablation study on the aggregated dataset on which we finetune the arbitrary keypoints prediction models, by removing one of the dataset for each finetuning run.\\

We perform this ablation study on both RICH dataset and OpenCapBench. The results are summarized in tables \ref{table:ablation_results_rich} and \ref{table:ablation_results_ocb}.

\begin{table}[h]
\centering
\scalebox{0.7}{
\setlength\tabcolsep{10pt}
\begin{tabular}{@{}lccc@{}}\toprule

& \multicolumn{3}{c}{PCK Precision(\textbf{$\uparrow$})}\\ \cmidrule{2-4}
& @0.05 & @0.1 & @0.2\\ \midrule
\multicolumn{4}{l}{\textit{Full agg. dataset}}\\
\phantom{ab} HRNet-W48& 0.89 & 0.96 & 0.98\\
\phantom{ab} ViTPose-B & 0.86 & 0.94 & 0.97\\
\phantom{ab} ViTPose-H & \textbf{0.90} & \textbf{0.97} & \textbf{0.99}\\
\multicolumn{4}{l}{\textit{Without COCO Dataset}}\\
\phantom{ab} HRNet-W48 & 0.87  \textcolor{red}{(-0.02)} & 0.94 \textcolor{red}{(-0.02)} & 0.96 \textcolor{red}{(-0.02)}\\
\phantom{ab} ViTPose-B & 0.83 \textcolor{red}{(-0.03)} & 0.91 \textcolor{red}{(-0.03)} & 0.95 \textcolor{red}{(-0.02)}\\
\phantom{ab} ViTPose-H & {0.90} (0.00) & 0.96 \textcolor{red}{(-0.01)} & 0.98 \textcolor{red}{(-0.01)}\\
\multicolumn{4}{l}{\textit{Without BEDLAM Dataset}}\\
\phantom{ab} HRNet-W48 & 0.87 \textcolor{red}{(-0.02)} & 0.95 \textcolor{red}{(-0.01)} & 0.97 \textcolor{red}{(-0.01)}\\\phantom{ab} ViTPose-B & 0.85 \textcolor{red}{(-0.01)} & 0.93 \textcolor{red}{(-0.01)} & 0.97 (0.00)\\
\phantom{ab} ViTPose-H & 0.89 \textcolor{red}{(-0.01)} & 0.96 \textcolor{red}{(-0.01)} & 0.98 \textcolor{red}{(-0.01)}\\
\multicolumn{4}{l}{\textit{Without Infinity Data}}\\
\phantom{ab} HRNet-W48 & 0.89 (0.00) & 0.96 (0.00) & 0.98 (0.00)\\
\phantom{ab} ViTPose-B+DARK & 0.86 (0.00) & 0.94 (0.00) & 0.97 (0.00)\\
\phantom{ab} ViTPose-H+DARK & 0.90 (0.00) & 0.97 (0.00) & 0.99 (0.00)\\
\multicolumn{4}{l}{\textit{Without 3DPW Dataset}}\\
\phantom{ab} HRNet-W48 & 0.90 \textcolor{green}{(+0.01)} & 0.97 \textcolor{green}{(+0.01)} & 0.98 (0.00) \\\phantom{ab} ViTPose-B & 0.88  \textcolor{green}{(+0.2)} & 0.95  \textcolor{green}{(+0.01)} & 0.97 (0.00)\\
\phantom{ab} ViTPose-H & 0.91 \textcolor{green}{(+0.01)}& 0.97 (0.00) & 0.99 (0.00)\\
\bottomrule
\end{tabular}
}
\vspace{2mm}
\caption{Ablation Study results on RICH\cite{rich} test set using SynthPose. Decrease/increase in performance over baseline are indicated in red/green. The study indicates slight negative impact when removing COCO and BEDLAM, and slight positive impact when removing 3DPW from the training set.}
\label{table:ablation_results_rich}
\vspace{-5mm}
\end{table}

\begin{table*}[t]
\centering
    \scalebox{0.7}{
\ra{0.8}
\setlength\tabcolsep{2pt}
\begin{tabular}{@{}lcccccccccccc@{}}\toprule
\multicolumn{0}{l}{Joint angles RMSE (\textbf{$\downarrow$})} & \multicolumn{3}{c}{Pelvis} & \multicolumn{3}{c}{Hip} & Knee & Ankle & Subtalar & \multicolumn{3}{c}{Lumbar} \\ \cmidrule{2-4} \cmidrule{5-7} \cmidrule{11-13} & Tilt & Rotation & List & Flex & Add & Rot & & & & Ext & Bend & Rot \\ \midrule
\multicolumn{5}{l}{\textit{Full agg. dataset}}\\
\phantom{ab} HRNet-W48 & 5.9 & 2.9 & 3.5 & 8.9 & 4.9 & 7.3 & 8.9 & 8.7 & 9.5 & 7.5 & 4.7 & 8.3\\
\phantom{ab} ViTPose-B  & 5.2 & 3.3 & 3.6 & 8.3 & 5.3 & 8 & 8.6 & 8.7 & 9.3 & 6.6 & 5.1 & 10\\
\phantom{ab} ViTPose-H & 5.1 & 2.8 & 3.4 & 8.3 & 4.9 & 7.3 & 8.3 & 7.6 & 9.1 & 6.2 & 4.6 & 8.7\\
\multicolumn{5}{l}{\textit{Without COCO dataset}} \\
\phantom{ab}{HRNet-W48} & (-0.1) & (+0.1) & (+0.1) & (+0.2) & (+0.1) & (+0.2) & \textcolor{red}{(+1.1)} & (+0.3) & (+0.3) & (-0.1) & (+0.1) & (-0.1) \\
\phantom{ab}{ViTPose-B} & (+0.4) & (0.0) & (+0.2) & \textcolor{red}{(+1.2)} & (+0.2) & (-0.1) & (+0.6) & (+0.7) & (+0.4) & (+0.3) & (+0.1) & (0.0) \\
\phantom{ab}{ViTPose-H} & (-0.1) & (0.0) & (0.0) & (+0.1) & (0.0) & (-0.4) & (+0.3) & (0.0) & (0.0) & (-0.1) & (0.0) & (-0.2) \\
\multicolumn{5}{l}{\textit{Without BEDLAM Dataset}} \\
\phantom{ab}{HRNet-W48} & (-0.4) & (+0.1) & (+0.3) & (+0.4) & (+0.1) & (+0.3) & (+0.4) & (0.0) & (-0.1) & (+0.6) & (+0.2) & (+0.3) \\
\phantom{ab}{ViTPose-B} & (0.0) & (0.0) & (+0.2) & (+0.6) & (+0.1) & (+0.4) & (+0.4) & (+0.8) & (+0.4) & (+0.8) & (+0.1) & (-0.5) \\
\phantom{ab}{ViTPose-H} & (-0.5) & (+0.1) & (+0.1) & (-0.1) & (+0.0) & (-0.3) & (+0.1) & (+0.3) & (0.0) & (+0.1) & (+0.1) & (+0.1) \\
\multicolumn{5}{l}{\textit{Without Infinity Data}} \\
\phantom{ab}{HRNet-W48} & (+0.4) & (+0.1) & (+0.1) & (+0.7) & (+0.3) & (+0.8) & (+0.6) & \textcolor{red}{(+1.0)} & (+0.6) & \textcolor{red}{(+1.1)} & (+0.0) & (+0.2) \\
\phantom{ab}{ViTPose-B} & \textcolor{red}{(+2.9)} & (+0.2) & (0.0) & \textcolor{red}{(+2.8)} & (+0.4) & (+0.8) & (-0.1) & (+0.4) & (+0.5) & \textcolor{red}{(+2.3)} & (+0.1) & (-0.1) \\
\phantom{ab}{ViTPose-H} & \textcolor{red}{(+2.2)} & (+0.2) & (-0.1) & \textcolor{red}{(+2.2)} & (+0.3) & (+0.5) & (+0.0) & (+0.6) & (+0.3) & \textcolor{red}{(+2.3)} & (0.0) & (+0.2) \\
\multicolumn{5}{l}{\textit{Without 3DPW Dataset}} \\
\phantom{ab}{HRNet-W48} & (-0.6) & (-0.1) & (+0.2) & (+0.1) & (+0.0) & (+0.6) & (+0.2) & (-0.1) & \textcolor{red}{(+1.1)} & \textcolor{green}{(-1.1)} & (+0.1) & (-0.5) \\
\phantom{ab}{ViTPose-B} & (0.0) & (0.0) & (+0.2) & (+0.4) & (+0.1) & (+0.1) & (+0.5) & (-0.2) & (+0.5) & (+0.7) & (+0.1) & (-0.2) \\
\phantom{ab}{ViTPose-H} & (+0.3) & (0.0) & (+0.2) & (+1.0) & (+0.0) & (+0.2) & (+0.8) & (-0.2) & \textcolor{red}{(+1.0)} & (+0.6) & (+0.0) & (-0.7) \\
\bottomrule
\vspace{2mm}

\end{tabular}}
\caption{Ablation study on OpenCapBench using SynthPose. Significant (>1.0) decrease/increase in performance over baseline are highlighted in red/green. The study shows the importance of Infinity data when it comes to prediciting accurate kinematics.}
\label{table:ablation_results_ocb}
\vspace{-5mm}

\end{table*}

In comparing the effects of various datasets on model performance, \ocb offers detailed insights that are not as evident in the RICH ablation study. 

The RICH ablation study, using PCK metric at different precision levels, shows no or very slight decrease in performance with the exclusion of each dataset, except without 3DPW, which appears to increase the models' performance. 

\ocb, on the other hand, provides a more detailed perspective, particularly highlighting the importance of the Infinity dataset for enhancing predictions on specific anatomical features such as Pelvis Tilt, Hip Flexion, and Lumbar Extension, potentially due to its focus on exercise-related data. It also reveals that while 3DPW may negatively impact some metrics, it is crucial for improving the Subtalar metric, which we hypothesize is due to the fact that 3DPW addresses the lack of subjects wearing shoes in the other datasets of the aggregated training set. This demonstrates \ocb's ability to offer nuanced insights that traditional pose estimation benchmarks cannot provide, into how different datasets uniquely contribute to model performance on biomechanical relevant metrics. 

\section{Discussions}
\label{sec:discussions}

OpenCapBench represents a step towards integrating kinematics and pose estimation, while introducing SynthPose, a method for estimating arbitrary keypoints which benefits both fields. This approach yields detailed insights into the performance of pose estimation models and the importance of diverse and comprehensive training data in refining these models.

Despite the benefits of OpenCapBench, the current dataset diversity within OpenCapBench currently lacks breadth in terms of subject variety, environmental settings, and the range of activities covered, which will be a focus of future work. Integrating additional datasets which use MoCap as ground truth such as MoYo \cite{moyo} or PSU-TMM100 \cite{psutmm100} may extend the benchmark's applicability and relevance across broader kinematic studies. 

At present, \ocb primarily focuses on lower body kinematics. Adding upper body kinematics and including upper limb assessments could help characterize more holistic view of human motion. 

While we focused on 2D pose estimation, we envision that Synthpose could benefit monocular 3D models as well, since they similarly suffer from keypoint scarcity as 2D models. We hope to work on such problems in the future or encourage the community to do so.

Finally, the open-source aspect and the versatility of OpenCapBench presents an opportunity for the community to engage with it through other pose estimation tasks such as 3D keypoint estimation and temporal predictions, as well as testing different subsets of keypoints and experimenting with new setups for inverse kinematics. This aspect encourages a collaborative approach, inviting contributions that could further the field of computer vision and kinematic analysis through the use of OpenCapBench.


\clearpage

\section*{Supplementary materials}

\subsection*{Finetuning details}

All SynthPose models were finetuned using the mmpose framework \cite{mmpose2020}, for 30 epochs each. The fine-tuning process utilized the Adam optimizer \cite{adam} and a multi-step learning rate scheduler. The learning rate started at \num{5e-3} and was reduced by a factor of 10 at epochs 15 and 20. The length of an epoch was determined by the size of the largest dataset. Throughout finetuning, we cycled through datasets using random index permutations for each cycle, ensuring an equal distribution of training samples from each dataset within each epoch and each batch when possible. For example, if a training batch is of size 4 and the aggregated dataset contains 4 datasets, each batch contains one sample from each dataset.

\subsection*{Importance of pretraining}
Although we briefly discussed the importance of utilizing weights from models trained on popular datasets like COCO to finetune SynthPose models in the main paper, we present a quantitative ablation study below.

\begin{table}[ht]
\centering
    \scalebox{0.92}{
\ra{0.8}
\setlength\tabcolsep{10pt}
\begin{tabular}{@{}lccc@{}}\toprule

& \multicolumn{3}{c}{PCK Precision(\textbf{$\uparrow$})}\\ \cmidrule{2-4}
& @0.05 & @0.10 & @0.20\\ \midrule
\multicolumn{4}{l}{\textit{SynthPose trained from scratch}} \vspace{2pt}\\
\phantom{ab} HRNet-W48 \cite{hrnetDark} & 0.843 & 0.929 & 0.967\\
\phantom{ab} ViTPose-B \cite{vitpose} &  0.738 & 0.851 & 0.912\\
\phantom{ab} ViTPose-H \cite{vitpose} & 0.644 & 0.769 & 0.853\\
\multicolumn{4}{l}{\textit{SynthPose finetuned from COCO pretrained weights}}\vspace{2pt}\\
\phantom{ab} HRNet-W48 \cite{hrnetDark} & 0.892 & 0.958 & 0.982\\
\phantom{ab} ViTPose-B \cite{vitpose} & 0.859 & 0.941 & 0.971\\
\phantom{ab} ViTPose-H \cite{vitpose} & \textbf{0.903} & \textbf{0.966} & \textbf{0.985}\\
\bottomrule
\end{tabular}
}
\vspace{2mm}
\caption{
Ablation study on the RICH \cite{rich} test set, illustrating the importance of leveraging pretrained weights to fine-tune SynthPose models. Additionally, we observe that the significance of using pretrained weights increases with the number of parameters in a model.}
\label{table:ablation_pretrain}
\vspace{-5mm}
\end{table}

\newpage

\subsection*{Visualizations}
\label{sec:visualizations}

\begin{figure}[h]
\centering
\includegraphics[width=0.7\linewidth]{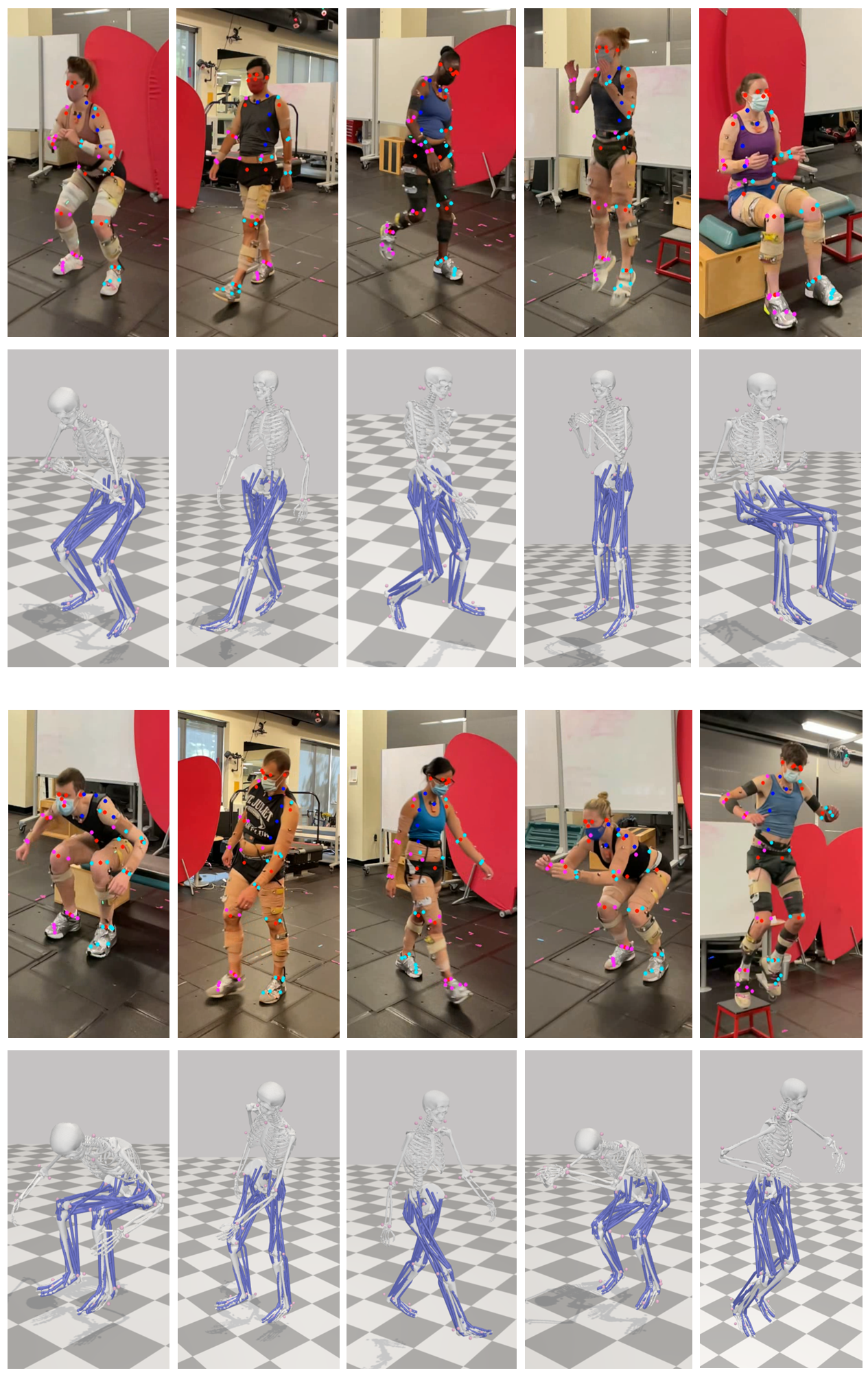}
\caption{Samples of Synthpose MoCap marker predictions on each subject of the OpenCap dataset, with their corresponding OpenSim kinematics output. The pink markers represent anatomical markers on the right side of the body, the cyan markers represent those on the left side, the blue markers represent markers in the center of the body, and the red markers represent markers in the COCO format. The SynthPose model used is HRNet48 fine-tuned on the entire aggregated dataset to predict the subset of markers defined in the main submission.}
\label{fig:viz}
\vspace{-20mm}
\end{figure}

\clearpage

\newpage
%
%
\bibliographystyle{splncs04}
\bibliography{main}

\begin{thebibliography}{10}
\providecommand{\url}[1]{\texttt{#1}}
\providecommand{\urlprefix}{URL }
\providecommand{\doi}[1]{https://doi.org/#1}

\bibitem{Infinity}
{Infinity AI VisionFit API}. \url{https://infinity.ai/visionfit}

\bibitem{kine1984}
An, K.N.: Kinematic analysis of human movement. Annals of Biomedical Engineering  \textbf{12}(6),  585--597 (Nov 1984)

\bibitem{posetrack}
Andriluka, M., Iqbal, U., Insafutdinov, E., Pishchulin, L., Milan, A., Gall, J., Schiele, B.: Posetrack: A benchmark for human pose estimation and tracking (2018)

\bibitem{MPII}
Andriluka, M., Pishchulin, L., Gehler, P., Schiele, B.: 2d human pose estimation: New benchmark and state of the art analysis. In: IEEE Conference on Computer Vision and Pattern Recognition (CVPR) (June 2014)

\bibitem{ICsynth}
Azizi, S., Kornblith, S., Saharia, C., Norouzi, M., Fleet, D.J.: Synthetic data from diffusion models improves imagenet classification (2023)

\bibitem{moncularKinematics}
Bittner, M., Yang, W.T., Zhang, X., Seth, A., van Gemert, J., van~der Helm, F.C.T.: Towards single camera human 3d-kinematics. Sensors  \textbf{23}(1) (2023). \doi{10.3390/s23010341}, \url{https://www.mdpi.com/1424-8220/23/1/341}

\bibitem{BEDLAM}
Black, M.J., Patel, P., Tesch, J., Yang, J.: {BEDLAM}: A synthetic dataset of bodies exhibiting detailed lifelike animated motion. In: Proceedings IEEE/CVF Conf.~on Computer Vision and Pattern Recognition (CVPR). pp. 8726--8737 (Jun 2023)

\bibitem{OpenPose}
Cao, Z., Hidalgo, G., Simon, T., Wei, S.E., Sheikh, Y.: Openpose: Realtime multi-person 2d pose estimation using part affinity fields (2019)

\bibitem{healthsynth}
Chambon, P., Bluethgen, C., Delbrouck, J.B., der Sluijs, R.V., Połacin, M., Chaves, J.M.Z., Abraham, T.M., Purohit, S., Langlotz, C.P., Chaudhari, A.: Roentgen: Vision-language foundation model for chest x-ray generation (2022)

\bibitem{kine2014}
Chèze, L.: Some Clinical Applications, chap.~5, pp. 73--101. John Wiley \& Sons, Ltd (2014). \doi{https://doi.org/10.1002/9781119058144.ch5}, \url{https://onlinelibrary.wiley.com/doi/abs/10.1002/9781119058144.ch5}

\bibitem{kinect}
Clark, R., Pua, Y.H., Oliveira, C.C., Bower, K., Thilarajah, S., McGaw, R., Hasanki, K., Mentiplay, B.: Reliability and concurrent validity of the microsoft xbox one kinect for assessment of standing balance and postural control. Gait \& Posture  \textbf{42} (04 2015). \doi{10.1016/j.gaitpost.2015.03.005}

\bibitem{mmpose2020}
Contributors, M.: Openmmlab pose estimation toolbox and benchmark. \url{https://github.com/open-mmlab/mmpose} (2020)

\bibitem{knee_dof}
Delp, S.L., Loan, J.P., Hoy, M.G., Zajac, F.E., Topp, E.L., Rosen, J.M.: An interactive graphics-based model of the lower extremity to study orthopaedic surgical procedures. IEEE Trans Biomed Eng  \textbf{37}(8),  757--767 (Aug 1990)

\bibitem{kineAcc}
Gholami, M., Napier, C., Menon, C.: Estimating lower extremity running gait kinematics with a single accelerometer: A deep learning approach. Sensors  \textbf{20}(10) (2020). \doi{10.3390/s20102939}, \url{https://www.mdpi.com/1424-8220/20/10/2939}

\bibitem{movi}
Ghorbani, S., Mahdaviani, K., Thaler, A., Kording, K., Cook, D.J., Blohm, G., Troje, N.F.: Movi: A large multi-purpose human motion and video dataset. PLOS ONE  \textbf{16}(6),  e0253157 (Jun 2021). \doi{10.1371/journal.pone.0253157}, \url{http://dx.doi.org/10.1371/journal.pone.0253157}

\bibitem{rich}
Huang, C.H.P., Yi, H., H{\"o}schle, M., Safroshkin, M., Alexiadis, T., Polikovsky, S., Scharstein, D., Black, M.J.: Capturing and inferring dense full-body human-scene contact. In: Proceedings IEEE/CVF Conf.~on Computer Vision and Pattern Recognition (CVPR). pp. 13274--13285 (Jun 2022)

\bibitem{humans36}
Ionescu, C., Papava, D., Olaru, V., Sminchisescu, C.: Human3.6m: Large scale datasets and predictive methods for 3d human sensing in natural environments. IEEE Transactions on Pattern Analysis and Machine Intelligence  \textbf{36}(7),  1325--1339 (2014). \doi{10.1109/TPAMI.2013.248}

\bibitem{handKinematics}
Isaac, J.H.R., Manivannan, M., Ravindran, B.: Corrective filter based on kinematics of human hand for pose estimation. Frontiers in Virtual Reality  \textbf{2} (2021). \doi{10.3389/frvir.2021.663618}, \url{https://www.frontiersin.org/articles/10.3389/frvir.2021.663618}

\bibitem{JHMDB}
Jhuang, H., Gall, J., Zuffi, S., Schmid, C., Black, M.J.: Towards understanding action recognition. In: International Conf. on Computer Vision (ICCV). pp. 3192--3199 (Dec 2013)

\bibitem{poseAnatomicalConstraints}
Ji, Z., Wang, Z., Zhang, M., Chen, Y., Qian, Y.: 2d human pose estimation with explicit anatomical keypoints structure constraints (2022)

\bibitem{cmupanoptic}
Joo, H., Simon, T., Li, X., Liu, H., Tan, L., Gui, L., Banerjee, S., Godisart, T.S., Nabbe, B., Matthews, I., Kanade, T., Nobuhara, S., Sheikh, Y.: Panoptic studio: A massively multiview system for social interaction capture. IEEE Transactions on Pattern Analysis and Machine Intelligence  (2017)

\bibitem{SKEL}
Keller, M., Werling, K., Shin, S., Delp, S., Pujades, S., C.~Karen, L., Black, M.J.: From skin to skeleton: Towards biomechanically accurate 3d digital humans. In: ACM ToG, Proc.~SIGGRAPH Asia. vol.~42 (Dec 2023)

\bibitem{single_cam_mov_analysis}
KidziAski, A., Yang, B., Hicks, J., Rajagopal, A., Delp, S., Schwartz, M.: Deep neural networks enable quantitative movement analysis using single-camera videos. Nature Communications  \textbf{11}, ~4054 (08 2020). \doi{10.1038/s41467-020-17807-z}

\bibitem{adam}
Kingma, D.P., Ba, J.: Adam: A method for stochastic optimization (2017)

\bibitem{sportInjuries}
Krosshaug, T., Nakamae, A., Boden, B., Engebretsen, L., Smith, G., Slauterbeck, J., Hewett, T., Bahr, R.: Mechanisms of anterior cruciate ligament injury in basketball: Video analysis of 39 cases. The American journal of sports medicine  \textbf{35},  359--67 (04 2007). \doi{10.1177/0363546506293899}

\bibitem{cliff}
Li, Z., Liu, J., Zhang, Z., Xu, S., Yan, Y.: Cliff: Carrying location information in full frames into human pose and shape estimation (2022)

\bibitem{COCO}
Lin, T.Y., Maire, M., Belongie, S., Bourdev, L., Girshick, R., Hays, J., Perona, P., Ramanan, D., Zitnick, C.L., Dollár, P.: Microsoft coco: Common objects in context (2015)

\bibitem{convnext}
Liu, Z., Mao, H., Wu, C.Y., Feichtenhofer, C., Darrell, T., Xie, S.: A convnet for the 2020s. Proceedings of the IEEE/CVF Conference on Computer Vision and Pattern Recognition (CVPR)  (2022)

\bibitem{SMPL:2015}
Loper, M., Mahmood, N., Romero, J., Pons-Moll, G., Black, M.J.: {SMPL}: A skinned multi-person linear model. ACM Trans. Graphics (Proc. SIGGRAPH Asia)  \textbf{34}(6),  248:1--248:16 (Oct 2015)

\bibitem{virtualMarkers}
Ma, X., Su, J., Wang, C., Zhu, W., Wang, Y.: 3d human mesh estimation from virtual markers (2023)

\bibitem{3DPW}
von Marcard, T., Henschel, R., Black, M., Rosenhahn, B., Pons-Moll, G.: Recovering accurate 3d human pose in the wild using imus and a moving camera. In: European Conference on Computer Vision (ECCV) (sep 2018)

\bibitem{3DjointsAcc}
Needham, L., Evans, M., Cosker, D.P., Wade, L., McGuigan, P.M., Bilzon, J.L., Colyer, S.L.: The accuracy of several pose estimation methods for 3d joint centre localisation. Scientific Reports  \textbf{11}(1),  20673 (Oct 2021). \doi{10.1038/s41598-021-00212-x}, \url{https://doi.org/10.1038/s41598-021-00212-x}

\bibitem{AGORA}
Patel, P., Huang, C.H.P., Tesch, J., Hoffmann, D.T., Tripathi, S., Black, M.J.: {AGORA}: Avatars in geography optimized for regression analysis. In: Proceedings IEEE/CVF Conf.~on Computer Vision and Pattern Recognition ({CVPR}) (Jun 2021)

\bibitem{SMPL-X:2019}
Pavlakos, G., Choutas, V., Ghorbani, N., Bolkart, T., Osman, A.A.A., Tzionas, D., Black, M.J.: Expressive body capture: {3D} hands, face, and body from a single image. In: Proceedings IEEE Conf. on Computer Vision and Pattern Recognition (CVPR). pp. 10975--10985 (2019)

\bibitem{Rajagopal}
Rajagopal, A., Dembia, C.L., DeMers, M.S., Delp, D.D., Hicks, J.L., Delp, S.L.: Full-body musculoskeletal model for muscle-driven simulation of human gait. IEEE Transactions on Biomedical Engineering  \textbf{63}(10),  2068--2079 (2016). \doi{10.1109/TBME.2016.2586891}

\bibitem{kineIMU}
Rapp, E., Shin, S., Thomsen, W., Ferber, R., Halilaj, E.: Estimation of kinematics from inertial measurement units using a combined deep learning and optimization framework. Journal of Biomechanics  \textbf{116},  110229 (01 2021). \doi{10.1016/j.jbiomech.2021.110229}

\bibitem{injuries}
Reinschmidt, C., van~den Bogert, A.J., Nigg, B.M., Lundberg, A., Murphy, N.: Effect of skin movement on the analysis of skeletal knee joint motion during running. Journal of biomechanics  \textbf{30 7},  729--32 (1997), \url{https://api.semanticscholar.org/CorpusID:38019421}

\bibitem{ResearchMethodsinBiomechanics}
Robertson, D., Caldwell, G., Hamill, J., Kamen, G., Whittlesey, S.: Research Methods in Biomechanics: Second edition (eBook) (11 2013)

\bibitem{psutmm100}
Scott, J.: Dynamic Stability Monitoring of Complex Human Motion Sequences via Precision Computer Vision. Ph.D. thesis (2022 2022), \url{https://www.proquest.com/dissertations-theses/dynamic-stability-monitoring-complex-human-motion/docview/2780914460/se-2}, copyright - Database copyright ProQuest LLC; ProQuest does not claim copyright in the individual underlying works; Last updated - 2023-11-10

\bibitem{biopose}
Scott, J., Ravichandran, B., Funk, C., Collins, R.T., Liu, Y.: From image to stability: Learning dynamics from human pose. In: Computer Vision--ECCV 2020: 16th European Conference, Glasgow, UK, August 23--28, 2020, Proceedings, Part XXIII 16. pp. 536--554. Springer (2020)

\bibitem{OpenSim}
Seth, A., Hicks, J.L., Uchida, T.K., Habib, A., Dembia, C.L., Dunne, J.J., Ong, C.F., DeMers, M.S., Rajagopal, A., Millard, M., Hamner, S.R., Arnold, E.M., Yong, J.R., Lakshmikanth, S.K., Sherman, M.A., Ku, J.P., Delp, S.L.: Opensim: Simulating musculoskeletal dynamics and neuromuscular control to study human and animal movement. PLOS Computational Biology  \textbf{14}(7),  1--20 (07 2018). \doi{10.1371/journal.pcbi.1006223}, \url{https://doi.org/10.1371/journal.pcbi.1006223}

\bibitem{joints_angles}
Tan, T., Gatti, A., Fan, B., Shea, K., Sherman, S., Uhlrich, S., Hicks, J., Delp, S., Shull, P., Chaudhari, A.: Towards out-of-lab anterior cruciate ligament injury prevention and rehabilitation assessment: A review of portable sensing approaches (10 2022). \doi{10.1101/2022.10.19.22281252}

\bibitem{moyo}
Tripathi, S., M{\"u}ller, L., Huang, C.H.P., Omid, T., Black, M.J., Tzionas, D.: {3D} human pose estimation via intuitive physics. In: Conference on Computer Vision and Pattern Recognition ({CVPR}). pp. 4713--4725 (2023), \url{https://ipman.is.tue.mpg.de}

\bibitem{OpenCap}
Uhlrich, S.D., Falisse, A., Kidzi{\'n}ski, {\L}., Muccini, J., Ko, M., Chaudhari, A.S., Hicks, J.L., Delp, S.L.: Opencap: 3d human movement dynamics from smartphone videos. bioRxiv  (2022). \doi{10.1101/2022.07.07.499061}, \url{https://www.biorxiv.org/content/early/2022/07/10/2022.07.07.499061}

\bibitem{SURREAL}
Varol, G., Romero, J., Martin, X., Mahmood, N., Black, M.J., Laptev, I., Schmid, C.: Learning from synthetic humans. In: CVPR (2017)

\bibitem{NLPsynth}
Wei, J., Huang, D., Lu, Y., Zhou, D., Le, Q.V.: Simple synthetic data reduces sycophancy in large language models (2023)

\bibitem{InfiniteForm}
Weitz, A., Colucci, L., Primas, S., Bent, B.: Infiniteform: A synthetic, minimal bias dataset for fitness applications (2021)

\bibitem{ISB}
Wu, G., Siegler, S., Allard, P., Kirtley, C., Leardini, A., Rosenbaum, D., Whittle, M., D’Lima, D.D., Cristofolini, L., Witte, H., Schmid, O., Stokes, I.: Isb recommendation on definitions of joint coordinate system of various joints for the reporting of human joint motion—part i: ankle, hip, and spine. Journal of Biomechanics  \textbf{35}(4),  543--548 (2002). \doi{https://doi.org/10.1016/S0021-9290(01)00222-6}, \url{https://www.sciencedirect.com/science/article/pii/S0021929001002226}

\bibitem{vitpose}
Xu, Y., Zhang, J., Zhang, Q., Tao, D.: Vi{TP}ose: Simple vision transformer baselines for human pose estimation. In: Advances in Neural Information Processing Systems (2022)

\bibitem{hrnetDark}
Zhang, F., Zhu, X., Dai, H., Ye, M., Zhu, C.: Distribution-aware coordinate representation for human pose estimation. In: IEEE/CVF Conference on Computer Vision and Pattern Recognition (CVPR) (June 2020)

\bibitem{poseSurvey}
Zheng, C., Wu, W., Chen, C., Yang, T., Zhu, S., Shen, J., Kehtarnavaz, N., Shah, M.: Deep learning-based human pose estimation: A survey (2023)

\end{thebibliography}
\end{document}